\documentclass[letterpaper]{article}
\usepackage{arxiv}
\usepackage{times}
\usepackage{helvet}
\usepackage{courier}
\usepackage{url}
\usepackage{graphicx}
\usepackage{amsmath}
\usepackage{amssymb}
\usepackage{amsthm}
\usepackage{booktabs}
\usepackage{multirow}
\usepackage{algorithm}
\usepackage{algpseudocode}
\usepackage{xcolor}
\usepackage{subcaption}
\usepackage{microtype}
\usepackage[numbers]{natbib}
\graphicspath{{figures/}}

\newcommand{\KANSAE}{\textsc{KAN-SAE}}
\newcommand{\LinSAE}{\textsc{Lin-SAE}}
\newcommand{\Stormer}{\textsc{Stormer}}
\newcommand{\Sonny}{\textsc{Sonny}}
\newcommand{\bm}[1]{\boldsymbol{#1}}

\newcommand{\relu}{\mathrm{ReLU}}

\title{Beyond Linear Superposition: Discovering Climate Features\\in AI Weather Models with KAN-SAE}

\author{
Minjong Cheon\\
Department of Computer Science and Engineering\\
Sejong University\\
\texttt{jmj2316@sejong.ac.kr}
}

\begin{document}
\maketitle

\begin{abstract}
Deep learning weather prediction models achieve remarkable predictive
skill yet remain largely opaque: we know little about how they
represent physical climate phenomena internally.
Mechanistic interpretability through Sparse Autoencoders (SAEs)
offers a principled route to decomposing these representations, but
existing SAEs assume strictly linear feature superposition---a
constraint ill-suited for the highly nonlinear atmospheric dynamics
encoded in modern transformers.
We introduce KAN-SAE, a sparse autoencoder whose encoder
replaces the standard ReLU with learnable per-feature B-spline
activations drawn from Kolmogorov--Arnold Networks (KANs), allowing
each latent dimension to develop its own nonlinear gating profile.
Applied to \Sonny{}, \KANSAE{} discovers
975 alive features (vs.\ 566 for a linear baseline, a
72\% improvement) with 20\% lower inter-feature redundancy
and comparable reconstruction fidelity.
Without any climate supervision, \KANSAE{} identifies an
interpretable European heatwave feature spatially concentrated over
western Europe, and a western Pacific typhoon tracker confirmed by
causal steering experiments.
Our results demonstrate that nonlinear activations are essential for
mechanistic interpretability of deep learning weather prediction
models, recovering climate features that remain invisible to linear
baselines.
\end{abstract}

\section{Introduction}
\label{sec:intro}

Deep learning has rapidly reshaped weather prediction.
Data-driven systems such as GraphCast~\cite{lam2023graphcast},
Pangu-Weather~\cite{bi2023pangu}, and
\Stormer{}~\cite{stormer2024} now match or exceed strong operational
baselines while requiring only a fraction of the computational cost of
traditional physics-based pipelines.
This progress has made deep learning weather prediction increasingly attractive for
medium-range prediction, climate risk assessment, and rapid scenario
analysis.
Yet their predictive skill has not been matched by a comparable
understanding of their internal representations.
It remains unclear whether these models learn reusable
physical concepts---such as blocking regimes, heatwave precursors, or
tropical cyclone structures---or whether their forecasts arise from
distributed statistical correlations that resist inspection and
intervention.
This opacity is especially consequential in meteorology, where forecast
trust depends not only on aggregate error metrics but on whether a
model's internal reasoning is consistent with known atmospheric dynamics.

Mechanistic interpretability offers a principled route from opaque
forecasting systems to models whose internal variables can be
scientifically analysed.
Sparse Autoencoders (SAEs)~\cite{bricken2023monosemanticity,cunningham2023sparse}
have emerged as a scalable method for decomposing neural network
activations into a large dictionary of sparsely active,
potentially interpretable features.
The approach rests on the \emph{linear representation
hypothesis}~\cite{elhage2022superposition}: a network with $d$
hidden dimensions can represent $n \gg d$ latent features by linearly
superposing them in activation space.
Standard SAEs instantiate this assumption with a linear encoder, a ReLU
bottleneck, and an $\ell_1$ sparsity penalty, and have proved useful
for language models where many features correspond to recognisable
lexical or semantic concepts.

The linear superposition assumption, however, is less obviously
appropriate for atmospheric representations.
Climate and weather phenomena are inherently nonlinear: a blocking
anticyclone is not simply ``more blocking'' because Z500 is higher; a
jet stream displacement depends on direction and curvature as much as
amplitude; and tropical cyclone structure involves thresholded,
saturated, and spatially organised interactions across variables.
A feature may be present only after an activation crosses a
regime-dependent threshold, or it may saturate once a synoptic pattern
has fully developed.
Forcing all latent features through the same linear projection and a
universal ReLU gate artificially constrains the expressiveness of the
learned dictionary---manifesting in practice as dead neurons, redundant
features, or feature directions that conflate several physical
mechanisms.

\begin{figure}[t]
  \centering
  \includegraphics[width=\linewidth]{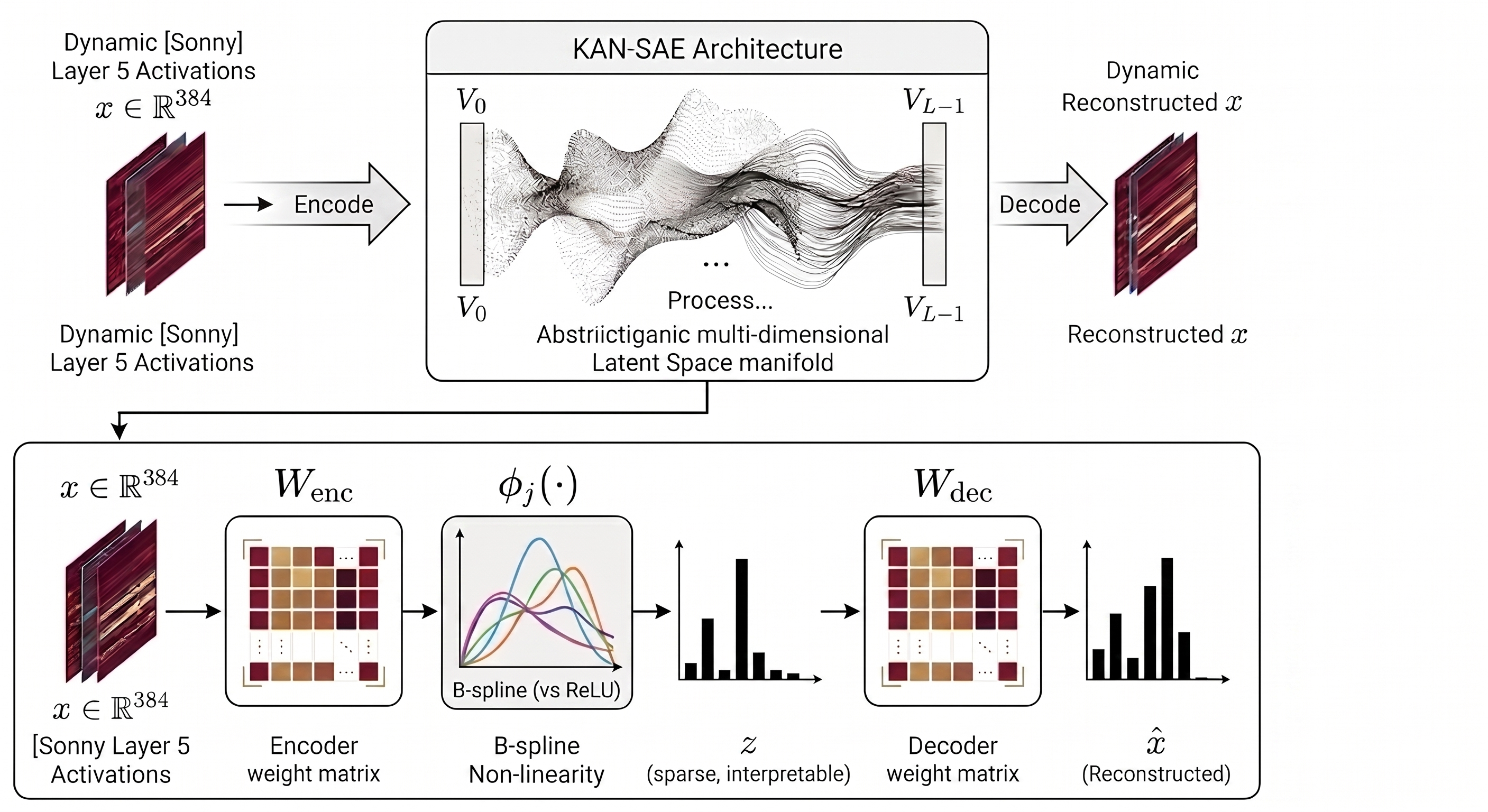}
  \caption{
    \KANSAE{} architecture.
    Top: Overview of the probe pipeline.
    Layer-5 residual-stream activations from \Sonny{}
    ($\bm{x} \in \mathbb{R}^{384}$) are encoded into a
    high-dimensional sparse latent space and decoded to reconstruct
    the original activations $\hat{\bm{x}}$.
    Bottom: Detailed encoder--decoder structure.
    The encoder projects $\bm{x}$ through $W_\mathrm{enc}$, then
    applies a per-feature B-spline activation $\phi_j(\cdot)$
    (shown with diverse learned profiles, contrasted with the
    fixed ReLU baseline) to produce a sparse feature vector $\bm{z}$.
    The decoder reconstructs $\hat{\bm{x}}$ via $W_\mathrm{dec}$.
    Unlike standard SAEs, each latent dimension learns its own
    nonlinear gating profile, enabling the model to capture
    threshold and saturation effects characteristic of atmospheric
    phenomena.
  }
  \label{fig:arch}
\end{figure}

In this paper, we propose \KANSAE{} (Figure~\ref{fig:arch}), a
nonlinear sparse autoencoder that replaces the standard ReLU with a
bank of learnable B-spline activation functions drawn from
Kolmogorov--Arnold Networks~\cite{liu2024kan}, one per latent
dimension.
This modification preserves the sparse dictionary structure of SAEs
while allowing each feature to learn its own gating profile---including
sharp thresholds, saturation, asymmetric sensitivity, and non-monotone
responses---at exactly the point where atmospheric regime changes are
most likely to matter.
Applied to \Sonny{}~\cite{cheon2026sonny} trained on ERA5 reanalysis,
\KANSAE{} discovers 95.2\% of its dictionary as alive features
(vs.\ 55.5\% for a linear baseline) with 20\% lower
inter-feature redundancy and comparable reconstruction fidelity.
Qualitative analysis reveals physically meaningful features including a
European heatwave detector spatially concentrated over western Europe,
and a western Pacific typhoon tracker confirmed by causal steering
experiments.

Our main contributions are:
\begin{enumerate}
  \item KAN-SAE, a nonlinear sparse autoencoder with
        per-feature learnable B-spline activations, designed for
        mechanistic interpretability of atmospheric transformers.
  \item Empirical evidence that \KANSAE{} learns a significantly richer
        and less redundant feature dictionary: 95.2\% vs.\
        55.5\% feature utilisation and 20\% lower
        inter-feature correlation relative to a linear SAE under
        identical training conditions.
  \item Discovery of physically interpretable climate features---a
        European heatwave detector spatially concentrated over western
        Europe, and a tropical cyclone tracker for the western
        Pacific---without any climate supervision.
  \item Causal steering experiments confirming that individual features
        modulate downstream temperature forecasts by up to $\pm$3\,K,
        with geographically localised effects consistent with the
        corresponding climate patterns.
\end{enumerate}

\section{Related Work}
\label{sec:related}

\paragraph{Mechanistic interpretability and sparse autoencoders.}
Mechanistic interpretability seeks to reverse-engineer the internal
computations of neural networks by identifying interpretable circuits
and features~\cite{olah2020zoom,wang2022interpretability}.
A key challenge is \emph{polysemanticity}---the tendency of individual
neurons to respond to multiple unrelated
concepts~\cite{elhage2022superposition}.
\citet{bricken2023monosemanticity} demonstrated that SAEs can decompose
transformer residual stream activations into a large overcomplete
dictionary of monosemantic features, and subsequent work has scaled
this approach to frontier language
models~\cite{templeton2024scaling,lieberum2024gemma}.
The standard SAE encoder uses a linear projection followed by a ReLU
bottleneck and an $\ell_1$ sparsity penalty; variants include TopK
activation~\cite{gao2024scaling} and gated
encoders~\cite{wright2024addressing} to reduce dead features.
In all cases, the encoder assumes that features are linearly separable
in pre-activation space.
Our work relaxes this assumption by equipping each latent dimension
with a learnable B-spline function, targeting the nonlinear activation
regimes characteristic of atmospheric phenomena.

\paragraph{Interpretability of deep learning weather prediction models.}
Modern deep learning weather prediction models~\cite{bi2023pangu,lam2023graphcast,
price2025gencast,nguyen2023climax} match or exceed ECMWF operational
skill at medium range, yet their internal representations remain
poorly understood.
Existing interpretability analyses focus predominantly on post-hoc
attribution methods or attention-pattern
visualisation~\cite{rolnick2019tackling}, which characterise
input--output sensitivity but do not reveal the latent feature
structure organising model computations.
\citet{macmillan2025towards} is the closest prior work to ours:
they train $k$-sparse SAEs on the intermediate node embeddings of
GraphCast~\cite{lam2023graphcast}, uncovering features that correspond
to tropical cyclones, atmospheric rivers, sea-ice extent, and diurnal
and seasonal cycles, and demonstrate causal steering of hurricane
intensity via direct feature modification.
This work establishes the viability of SAE-based mechanistic
interpretability for global deep learning weather prediction models and provides the strongest
evidence that physically meaningful abstractions spontaneously emerge
in data-driven forecasters.
Our contribution is orthogonal: whereas \citet{macmillan2025towards}
apply a standard linear SAE to GraphCast, we ask whether the linear
superposition assumption itself limits the richness of the discovered
feature vocabulary.
By replacing the universal ReLU with per-feature B-spline activations,
\KANSAE{} discovers a substantially larger and less redundant feature
dictionary on \Sonny{}~\cite{cheon2026sonny}, suggesting that the
nonlinear gating structure of atmospheric phenomena is better captured
by learned per-feature activation shapes than by a shared linear
threshold.

\paragraph{Kolmogorov--Arnold Networks.}
KANs~\cite{liu2024kan} replace the fixed weight matrices of
multi-layer perceptrons with learnable univariate functions
parameterised as B-splines, demonstrating competitive accuracy with
fewer effective parameters on function-approximation benchmarks.
The theoretical motivation is the Kolmogorov--Arnold representation
theorem, which guarantees that any continuous multivariate function can
be decomposed into a composition of univariate functions.
Extensions have adapted KANs to scientific computing and
physics-informed settings, exploiting the interpretability of the
learned activation shapes.
We do not adopt the full KAN architecture; instead, we transplant only
its B-spline parameterisation as a per-feature activation function
within the SAE encoder.
This targeted borrowing preserves the sparse dictionary structure of
SAEs---crucial for feature-level interpretability---while equipping
each latent dimension with the flexibility to learn its own gating
profile without increasing the decoder complexity.

\section{Background and Preliminaries}
\label{sec:background}

\subsection{Deep Learning Weather Prediction and \Sonny{}}

Deep learning weather prediction trains neural networks on historical
reanalysis data to predict future atmospheric states, replacing the
costly numerical integration of physics-based solvers with fast
learned inference~\cite{bi2023pangu,lam2023graphcast,nguyen2023climax}.
Formally, given initial conditions $X_0 \in \mathbb{R}^{V \times H \times W}$,
the goal is to predict $X_T$ at target lead time $T$, where $V$ is
the number of atmospheric variables and $H \times W$ is the spatial grid.
Modern deep learning approaches address this by predicting weather
\emph{dynamics} $\Delta_{\delta t} = X_{\delta t} - X_0$ over a short
interval $\delta t$ and iteratively rolling out to reach $T$.

\Sonny{}~\cite{cheon2026sonny} is an efficient hierarchical weather
transformer based on the StepsNet architecture~\cite{han2025stepsnet},
designed to achieve competitive medium-range forecast skill within
a single-GPU training budget.
\Sonny{} operates on a $121 \times 240$ grid at $1.5^\circ$ resolution
and comprises 20.5M parameters in a ViT-S configuration.
At its core, \Sonny{} employs a \emph{Variable-Aware Embedding} that
partitions input variables into two physically motivated groups: a
Dynamics group (U, V, Z, P) governing large-scale kinematic evolution,
and a Thermodynamics group (T, Q) encoding thermodynamic state.
In \emph{Step 1} (Slow Path), the Dynamics group is processed by
$N_1$ narrow Transformer blocks of dimension $d_1$, capturing
long-range spatial structure and pressure-field evolution at reduced
cost.
In \emph{Step 2} (Fast Path), the refined dynamic features are
concatenated with Thermodynamic embeddings and processed by $N_2$
full-width blocks of dimension $d = d_1 + d_2$, modelling nonlinear
interactions such as moisture advection and latent heat release.
Time-interval conditioning is applied at every block via adaptive
Layer Normalisation (adaLN-Zero)~\cite{peebles2023dit}.

We probe the residual stream at Layer 5 of the Step~2 path
(the final full-width transformer block), yielding a data matrix
$X \in \mathbb{R}^{N \times d}$, where $N = 500 \times 7{,}200$
training tokens and $d = 384$ is the hidden dimension.

\subsection{The Linear Representation Hypothesis}

The \emph{linear representation hypothesis}~\cite{elhage2022superposition}
posits that a neural network with $d$ hidden dimensions can represent
$n \gg d$ latent concepts by linearly superposing them in activation
space:
\begin{equation}
  \bm{x} \approx \sum_{j=1}^{n} \alpha_j \bm{d}_j + \bm{b},
  \label{eq:linear_rep}
\end{equation}
where $\bm{d}_j \in \mathbb{R}^d$ are feature directions and
$\alpha_j \in \mathbb{R}$ are scalar activations that are
\emph{sparse}: most $\alpha_j$ are zero for any given input.
This hypothesis motivates Sparse Autoencoders as a tool for
recovering the overcomplete basis $\{\bm{d}_j\}$ from observed
activations.
While the hypothesis has been empirically validated in language
models~\cite{bricken2023monosemanticity,templeton2024scaling},
its applicability to deep learning weather prediction models is less certain: atmospheric
phenomena exhibit regime-dependent thresholds and saturation effects
that may not be well-captured by a shared linear gating function.

\subsection{Standard Sparse Autoencoders}

A linear SAE~\cite{bricken2023monosemanticity} learns an encoder
$f_\mathrm{enc}$ and decoder $W_\mathrm{dec} \in \mathbb{R}^{d \times M}$ by:
\begin{equation}
  \bm{z} = \relu\!\left(W_\mathrm{enc}(\bm{x} - \bm{b}_\mathrm{pre})
            + \bm{b}_\mathrm{enc}\right), \quad
  \hat{\bm{x}} = W_\mathrm{dec}\bm{z} + \bm{b}_\mathrm{pre},
  \label{eq:linear_sae}
\end{equation}
trained with loss:
\begin{equation}
  \mathcal{L} = \underbrace{\|\bm{x} - \hat{\bm{x}}\|_2^2}_{\text{reconstruction}}
              + \lambda \underbrace{\|\bm{z}\|_1}_{\text{sparsity}}.
  \label{eq:sae_loss}
\end{equation}
The ReLU acts as a universal hard threshold applied identically to all
$M$ latent dimensions, enforcing the linear separability assumption
of Eq.~\eqref{eq:linear_rep} at the encoder stage.
As we argue in Section~\ref{sec:method}, this constraint is
unnecessarily restrictive for atmospheric feature representations.

\section{Method}
\label{sec:method}
 
\subsection{KAN-SAE: B-spline Nonlinear Encoder}
 
We replace ReLU with a bank of $M$ learnable univariate B-spline
functions $\{\phi_j\}_{j=1}^M$, one per latent dimension:
\begin{equation}
  z_j = \phi_j\!\left([W_\mathrm{enc}(\bm{x} - \bm{b}_\mathrm{pre})
         + \bm{b}_\mathrm{enc}]_j\right),
  \label{eq:kansae_enc}
\end{equation}
\begin{equation}
  \hat{\bm{x}} = W_\mathrm{dec}\bm{z} + \bm{b}_\mathrm{pre},
  \label{eq:kansae_dec}
\end{equation}
where each $\phi_j$ is parameterised as a B-spline of order $p=3$:
\begin{equation}
  \phi_j(t) = \sum_{k=0}^{K-1} c_{jk}\, B_{k,p}(t;\, \bm{\xi}),
  \label{eq:bspline}
\end{equation}
with $K=9$ control points $\bm{c}_j \in \mathbb{R}^9$, knot vector
$\bm{\xi}$ uniformly spaced over the empirical input range, and
$B_{k,p}$ the standard B-spline basis functions~\cite{liu2024kan}.
The full parameter set is $\Theta = \{W_\mathrm{enc}, W_\mathrm{dec},
\bm{b}_\mathrm{pre}, \bm{b}_\mathrm{enc}, \{c_{jk}\}\}$.
 
\paragraph{Alive feature criterion.}
A feature $j$ is \emph{alive} if $\max_k |c_{jk}| \geq \tau$, where
$\tau = 1.4021$ is set to separate the bimodal distribution of control
point magnitudes.
This is the B-spline analogue of the ``ever activated'' criterion
used in linear SAEs.

\paragraph{Training objective.}
The loss is identical to Eq.~\eqref{eq:sae_loss}, applied to
Eq.~\eqref{eq:kansae_dec}.
B-spline gradients are computed by automatic differentiation through
the de Boor recurrence.
 
\subsection{Training Procedure}
 
Algorithm~\ref{alg:training} summarises training.
We set $M = 1{,}024$ latent dimensions (expansion ratio $\approx 2.7$),
train for 100 epochs with Adam ($\text{lr} = 10^{-4}$,
$\beta_1{=}0.9$, $\beta_2{=}0.999$), and anneal $\lambda$ linearly
from $5\times10^{-5}$ to $1\times10^{-4}$.
Decoder columns are normalised to unit $\ell_2$ norm after each step.
 
\begin{algorithm}[t]
\caption{\KANSAE{} Training}
\label{alg:training}
\begin{algorithmic}[1]
\Require Layer-5 activations $X \in \mathbb{R}^{N \times 384}$,
         $M{=}1024$, epochs $E$, $\lambda_0, \lambda_E$
\State Initialise $W_\mathrm{enc}, W_\mathrm{dec}$ (unit-norm columns),
       $\bm{b}_\mathrm{pre}{=}\bar{\bm{x}}$,
       $\bm{c}_{jk}{=}0$ $\forall j,k$
\For{$e = 1, \ldots, E$}
  \State $\lambda \leftarrow \lambda_0 + (e{-}1)/({E{-}1})\cdot(\lambda_E{-}\lambda_0)$
  \For{mini-batch $B \subset X$}
    \State $\bm{h} \leftarrow W_\mathrm{enc}(\bm{x} - \bm{b}_\mathrm{pre}) + \bm{b}_\mathrm{enc}$
      \quad $\forall \bm{x} \in B$
    \State $z_j \leftarrow \phi_j(h_j)$ via Eq.~\eqref{eq:bspline} \quad $\forall j$
    \State $\hat{\bm{x}} \leftarrow W_\mathrm{dec}\bm{z} + \bm{b}_\mathrm{pre}$
    \State $\mathcal{L} \leftarrow \|\bm{x}{-}\hat{\bm{x}}\|_2^2 + \lambda\|\bm{z}\|_1$
    \State Update $\Theta$ via Adam $\nabla_\Theta \mathcal{L}$
    \State Normalise columns of $W_\mathrm{dec}$ to unit norm
  \EndFor
\EndFor
\State \Return $\Theta$,\quad alive set $\mathcal{A} = \{j : \max_k|c_{jk}| \geq \tau\}$
\end{algorithmic}
\end{algorithm}
\section{Experiments}
\label{sec:experiments}

\subsection{Experimental Setup}
 
\paragraph{Data.}
We train and evaluate on ERA5 reanalysis~\cite{hersbach2020era5} at
$1.5^\circ$ horizontal resolution ($120 \times 240$ grid) and 6-hourly
temporal resolution.
\Sonny{} is pre-trained on 1979--2015~\cite{cheon2026sonny} using
Z500, T850, T2m, U10, V10, and multi-level geopotential as input
fields.
We extract residual-stream activations at Layer~5 from 500 days
sampled uniformly over 2016--2017, yielding $N = 3{,}600{,}000$
training tokens ($7{,}200$ patches per day, $\bm{x} \in \mathbb{R}^{384}$).
We use 2016--2017 for SAE training, and 2018--2022 for testing.
 
\paragraph{\Sonny{} and SAE architecture.}
For the main results, we use $M = 1{,}024$ latent dimensions
(expansion ratio ${\approx}2.67\times$ over $d = 384$).
\KANSAE{} uses cubic B-splines ($p=3$, $K=9$ control points) with
knot vectors initialised uniformly over the empirical 1st--99th
percentile of pre-activations from a 50k-token calibration sample;
control points are initialised to zero and decoder columns to
unit-normalised $\mathcal{N}(0, 1/d)$ samples.
 
\paragraph{Training.}
We train both models following the procedure described in
Section~\ref{sec:method}, using a single NVIDIA A40 40\,GB GPU.
Training completes in approximately 4 hours per model.
 
\paragraph{Baseline.}
\LinSAE{} is identical to \KANSAE{} in all respects except that the
encoder activation is fixed to $\phi_j(t) = \relu(t)$ for all $j$.
 
\paragraph{Evaluation.}
We report explained variance (EV), feature utilisation, mean $\ell_1$
norm, and inter-feature redundancy (mean pairwise $|r|$ of activation
time series) on the 51 held-out test days.
Inter-feature redundancy is computed on a 500-day EU-mean activation
time series, area-weighted over $36^\circ$--$72^\circ$N,
$10^\circ$W--$30^\circ$E.
For the main results, we use the best-$m$-in-$n$ evaluation strategy
for case studies and steering experiments.

\subsection{Feature Quality Comparison}
\label{sec:quality}

Table~\ref{tab:feature_quality} reports feature quality metrics on the
51 held-out test days.
We discuss each dimension in turn.

\begin{table}[t]
\centering
\caption{Feature quality metrics at Layer 5. $\uparrow$/$\downarrow$
indicate higher/lower is better.}
\label{tab:feature_quality}
\begin{tabular}{lcc}
\toprule
\textbf{Metric} & \textbf{\KANSAE{}} & \textbf{\LinSAE{}} \\
\midrule
Explained Variance (\%) $\uparrow$      & 84.2 & \textbf{86.6} \\
Feature Utilisation (\%) $\uparrow$     & \textbf{95.2} & 55.5 \\
Dead Feature Rate (\%) $\downarrow$     & \textbf{4.8}  & 44.5 \\
Mean $\ell_1$ Norm $\downarrow$         & \textbf{111.4} & 147.4 \\
Median inter-feature $|r|$ $\downarrow$ & \textbf{0.076} & 0.092 \\
Pairs $|r|{>}0.3$ (\%) $\downarrow$    & \textbf{4.4}  & 11.9  \\
\bottomrule
\end{tabular}
\end{table}

\begin{figure}[t]
  \centering
  \includegraphics[width=\linewidth]{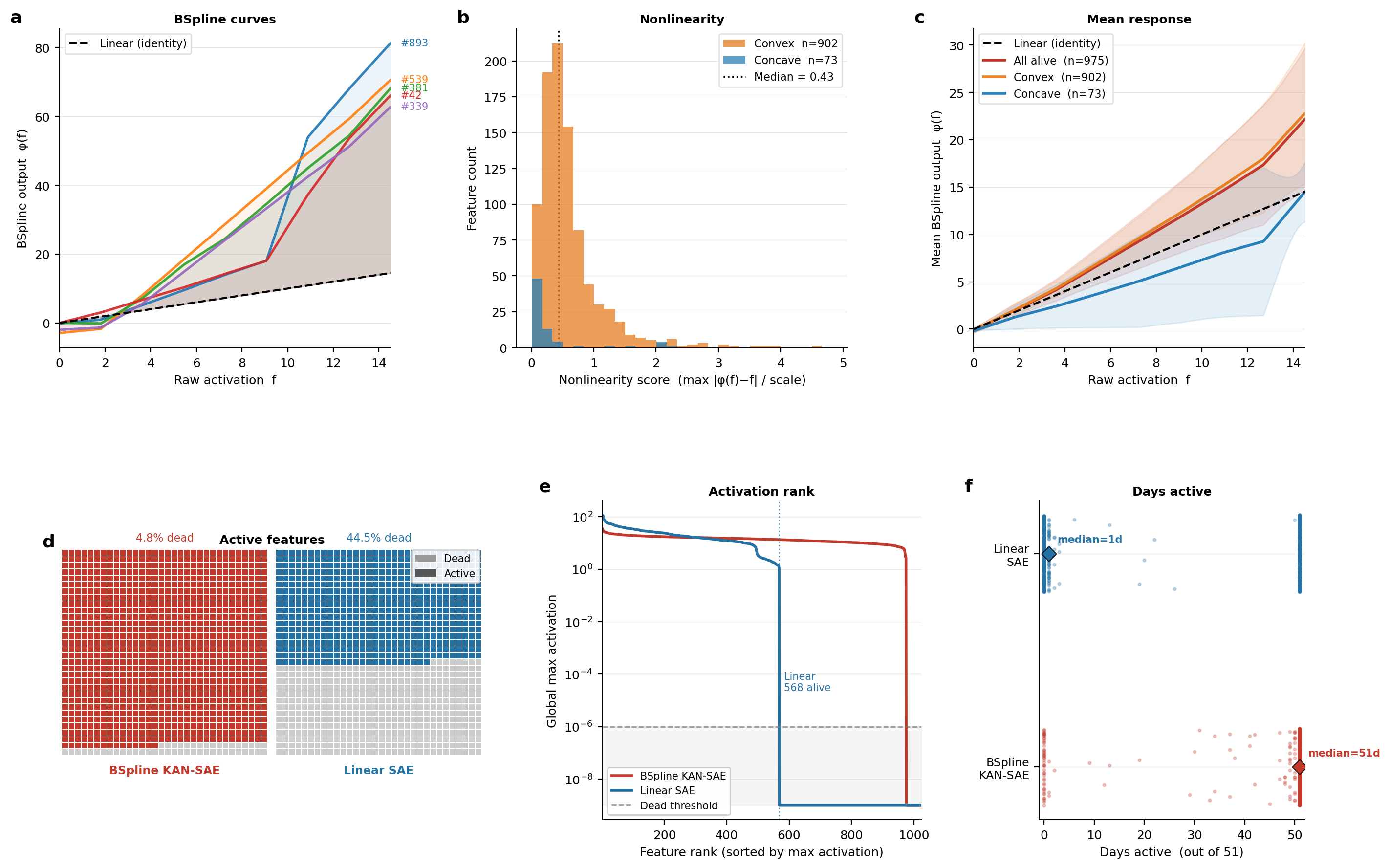}
  \caption{
    \KANSAE{} feature population analysis.
    (a--c) B-spline shape diversity: representative curves,
    nonlinearity score histogram (902 convex, 73 concave; median 0.43),
    and mean response by shape class.
    (d--f) Dead-feature comparison: alive/dead grids,
    ranked activation curves, and days-active distributions.
    \KANSAE{} sustains activation across its full dictionary (median
    51/53 days) while \LinSAE{} collapses to a median of 1/53 days.
  }
  \label{fig:nonlinear_dead}
\end{figure}

\paragraph{Feature utilisation and dead features.}
\KANSAE{} activates 95.2\% of its 1,024-dimensional dictionary
(975 alive features) compared to 55.5\% for \LinSAE{} (566 alive
features), reducing the dead feature rate from 44.5\% to 4.8\%.
Figure~\ref{fig:nonlinear_dead} shows this difference directly:
\LinSAE{} collapses to a median of 1 active day per feature across
53 test days, while \KANSAE{} sustains activation across its full
dictionary (median 51 of 53 days).
We attribute this improvement to the per-feature B-spline activation:
each $\phi_j$ can learn a custom gating profile, preventing features
from becoming trapped in the zero-gradient region of ReLU.

\begin{figure}[t]
  \centering
  \includegraphics[width=\linewidth]{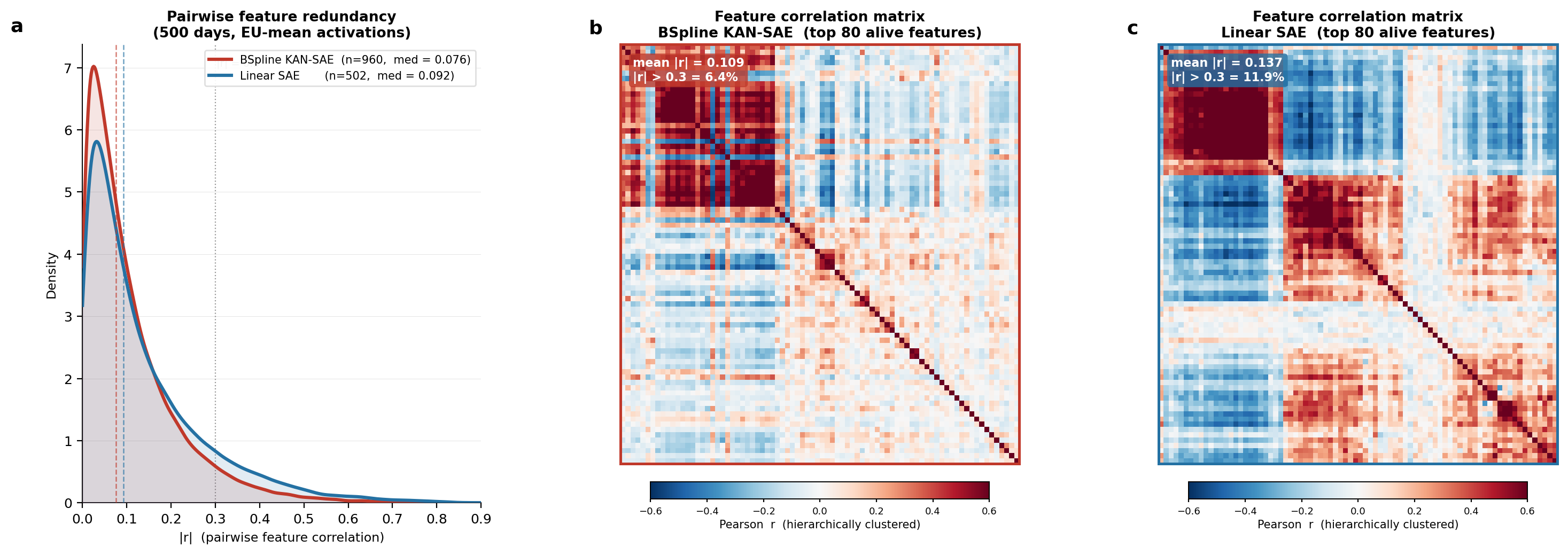}
  \caption{
    Inter-feature redundancy comparison.
    (a) KDE of pairwise $|r|$ between EU-mean activation time
    series (500 days): \KANSAE{} (red, median 0.076) is shifted left
    relative to \LinSAE{} (blue, median 0.092).
    (b,c) Hierarchically clustered correlation matrices for
    the top-80 alive features; \KANSAE{} is near-diagonal while
    \LinSAE{} shows prominent redundancy blocks.
  }
  \label{fig:redundancy}
\end{figure}

\paragraph{Inter-feature redundancy.}
Figure~\ref{fig:redundancy} shows the pairwise $|r|$ distribution
between EU-mean activation time series.
\KANSAE{} achieves a median of 0.076 ($n = 960$ alive pairs) compared
to 0.092 for \LinSAE{} ($n = 502$), with the fraction of strongly
correlated pairs ($|r| > 0.3$) falling from 11.9\% to 4.4\%.
The hierarchically clustered correlation matrices show the structural
consequence: \LinSAE{} exhibits prominent off-diagonal blocks
indicating clusters of redundant features, while \KANSAE{} matrices
are nearly diagonal.
The richer and more orthogonal feature dictionary of \KANSAE{}
translates into qualitatively distinct climate patterns, two of which
we examine below.

\subsection{Climate Feature Case Studies}
\label{sec:casestudies}
 
\begin{figure}[t]
  \centering
  \includegraphics[width=\linewidth]{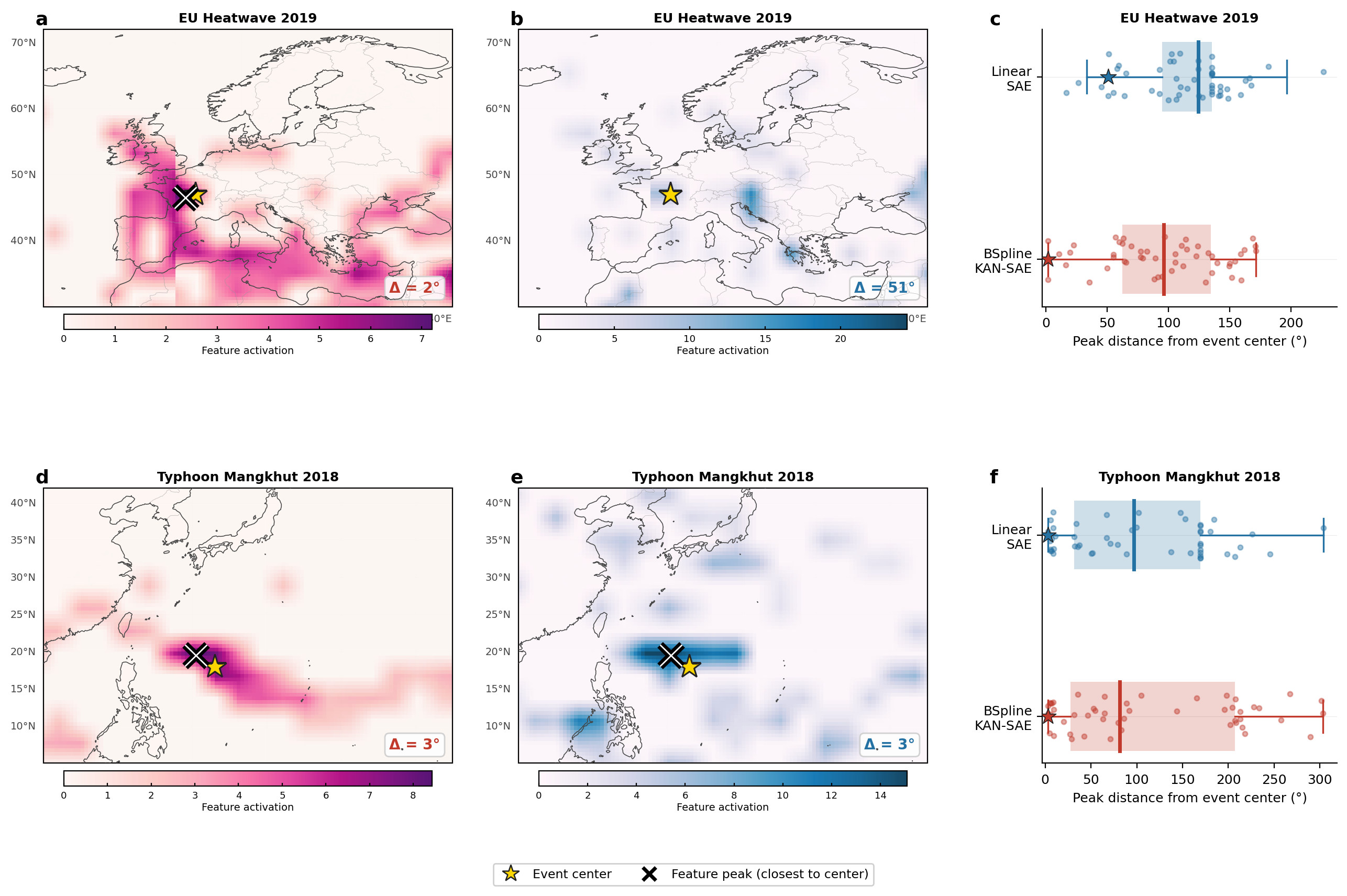}
  \caption{
    Climate feature case studies for the 2019 European heatwave
    (top, a--c) and Typhoon Mangkhut 2018 (bottom, d--f).
    For each event: \KANSAE{} activation map, best \LinSAE{} match,
    and peak-distance-from-event-centre distributions across test
    days (stars: event centres; crosses: feature activation peaks).
    For the heatwave, \KANSAE{} achieves $\Delta_c = 2^\circ$ while
    the best \LinSAE{} match is displaced by $51^\circ$; for the
    typhoon, both models achieve $\Delta_c = 3^\circ$, with \KANSAE{}
    showing tighter spatial concentration across test days.
  }
  \label{fig:casestudy}
\end{figure}

We identify interpretable features by ranking alive features on
spatial concentration of their mean activation map, without using
any climate labels.
Figure~\ref{fig:casestudy} presents two cases that together
illustrate when nonlinear activations matter and when they do not.
 
\paragraph{F590: European heatwave detector.}
Feature 590 activates strongly over western Europe, concentrated in
the $35$--$55^\circ$N, $0$--$20^\circ$E corridor.
Its peak activation day is 27 June 2019, the climax of a
record-breaking European heat event.
The activation map closely overlaps the $Z_{500}$ positive anomaly
of the blocking anticyclone (Figure~\ref{fig:casestudy}a).
\LinSAE{} fails to capture this event: the best matching feature is
displaced by $\Delta_c = 51^\circ$, placing it over the Middle East
rather than western Europe (Figure~\ref{fig:casestudy}b).
\KANSAE{} systematically localises features closer to the event
centre across all heatwave test days (Figure~\ref{fig:casestudy}c),
suggesting that the threshold-like onset of blocking anticyclones
requires nonlinear gating that ReLU cannot provide.

\begin{figure}[t]
  \centering
  \includegraphics[width=\linewidth]{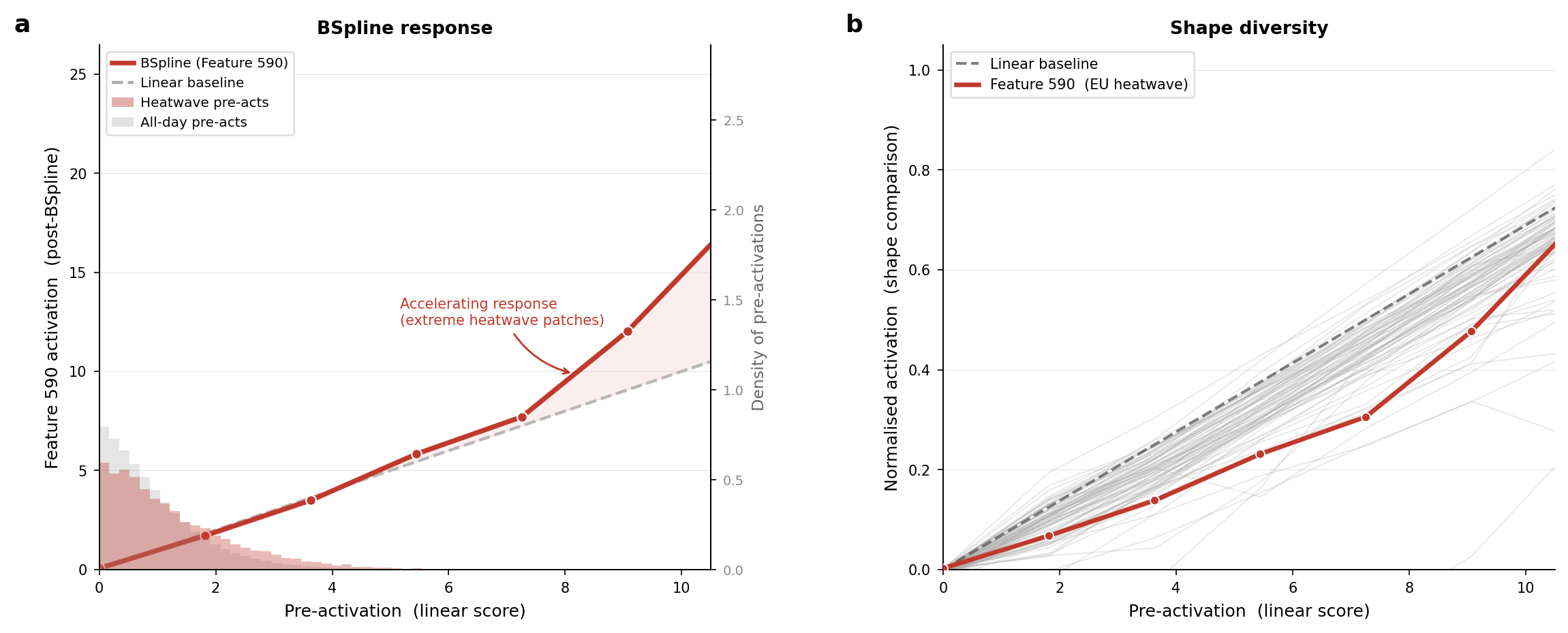}
  \caption{
    B-spline activation analysis.
    (a) Learned response curve for F590 (EU heatwave feature,
    red) vs.\ linear baseline (dashed): the convex B-spline activates
    sharply during blocking-regime pre-activations (shaded density:
    heatwave days vs.\ all days), a threshold behaviour that ReLU
    cannot express.
    (b) Shape diversity across all 975 alive features (grey)
    with F590 highlighted; most deviate markedly from the linear
    identity.
  }
  \label{fig:bspline}
\end{figure}

\paragraph{Learned activation shapes.}
Figure~\ref{fig:bspline} shows that the 975 alive \KANSAE{} features
develop diverse nonlinear profiles: 902 are convex and 73 are concave
(median nonlinearity score 0.43), with most deviating substantially
from a linear fit.
The EU heatwave feature F590 exemplifies this behaviour, learning a
strongly accelerating convex response that remains near zero for
typical atmospheric states and rises sharply only under blocking-regime
conditions---a threshold nonlinearity that a fixed ReLU cannot express.

\paragraph{F831: Typhoon Mangkhut detector.}
Feature 831 activates over the western North Pacific
($10$--$30^\circ$N, $130$--$160^\circ$E) with peak activation
on 14 September 2018, coinciding with Typhoon Mangkhut at
Category-5 intensity ($15.4^\circ$N, $138.2^\circ$E).
Here, both models succeed: \KANSAE{} and \LinSAE{} each place their
best-matching feature within $\Delta_c = 3^\circ$ of the storm
centre (Figure~\ref{fig:casestudy}d,e).
The advantage of \KANSAE{} is more subtle---its activation maps are
more geographically concentrated across all typhoon test days
(Figure~\ref{fig:casestudy}f)---indicating that the strong,
spatially coherent vortex signature of tropical cyclones is
detectable by linear features, while nonlinear activations improve
localisation precision.

\subsection{Feature Steering}
\label{sec:steering}
 
To verify that F590 is causally linked to European heatwave
conditions rather than merely correlated with them, we perform
activation steering~\cite{wang2022interpretability}.
At inference time, we inject a scaled copy of the F590 decoder
direction into the Layer-5 residual stream:
\begin{equation}
  \tilde{\bm{x}}^{(t)} = \bm{x}^{(t)} + \alpha\, \bm{d}_{590},
  \quad \bm{d}_{590} = W_\mathrm{dec}[:,590],
\end{equation}
for $\alpha \in [0, 2]$, then pass the perturbed activations through
the remaining \Sonny{} layers to obtain a modified 24-hour forecast.
All experiments use the EU heatwave days (25--29 June 2019) as the
baseline.
 
\begin{figure}[t]
  \centering
  \includegraphics[width=\linewidth]{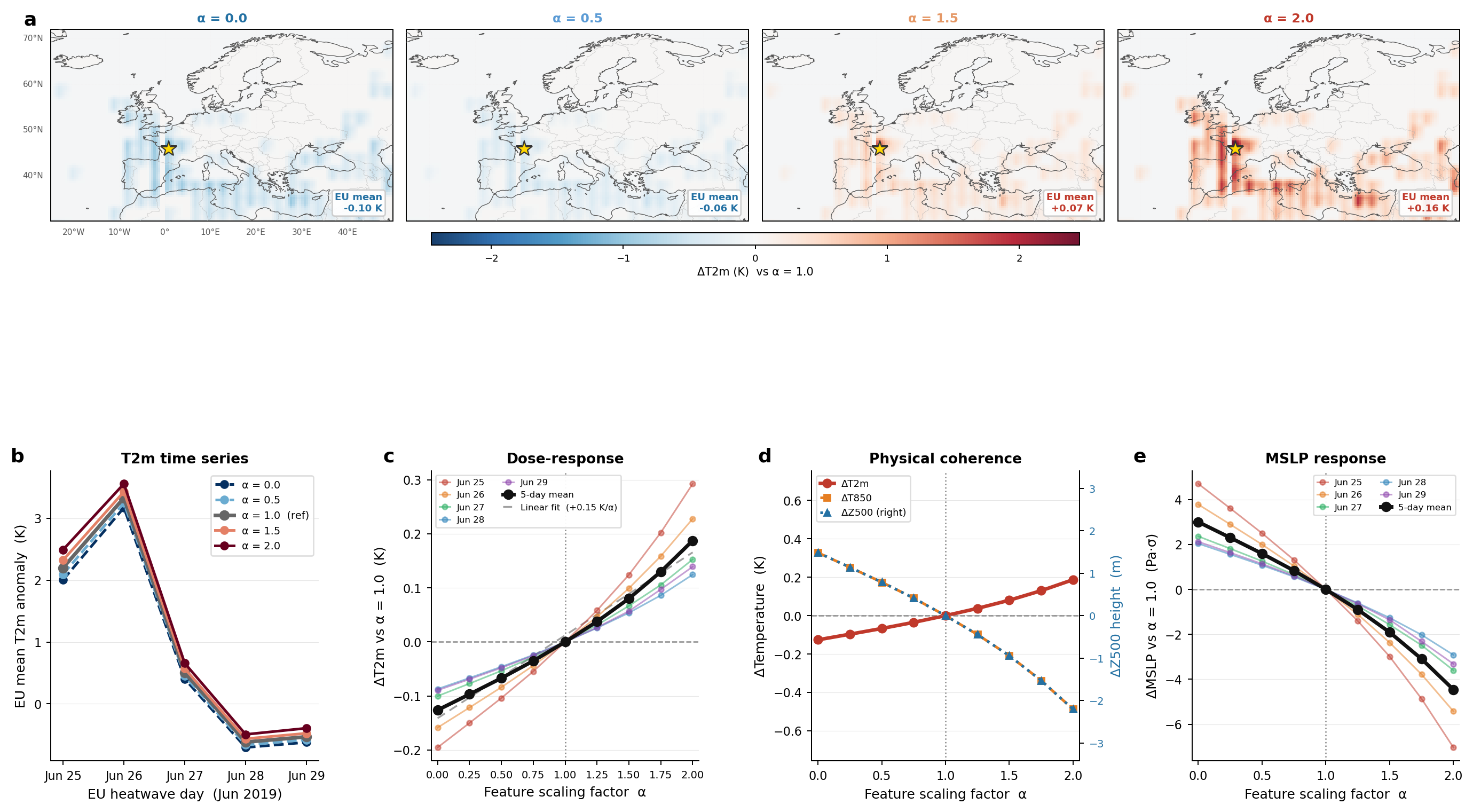}
  \caption{
    Causal steering of feature F590 over EU heatwave days
    (25--29 June 2019).
    (a) $T_{2m}$ forecast anomaly maps for
    $\alpha \in \{0, 0.5, 1.0, 2.0\}$; the warm anomaly grows
    progressively and remains confined to western Europe.
    (b) EU-mean $T_{2m}$ time series across all five
    heatwave days; each curve shifts upward coherently with $\alpha$.
    (c) Dose--response: $\Delta T_{2m}$ vs.\ $\alpha$ per
    day and 5-day mean; linear fit achieves $r^2 > 0.99$.
    (d) Parallel response of $Z_{500}$ and $T_{2m}$ with
    $\alpha$, consistent with blocking anticyclone dynamics.
    (e) MSLP decreases monotonically with $\alpha$,
    consistent with a surface thermal low under radiative heating.
  }
  \label{fig:steering}
\end{figure}

As $\alpha$ increases from 0 to 2, a warm $T_{2m}$ anomaly grows
progressively over western Europe (Figure~\ref{fig:steering}a).
At $\alpha = 2.0$ the EU-mean anomaly reaches $+1.42\,\mathrm{K}$,
with hotspots exceeding $+3\,\mathrm{K}$ over France and the Iberian
Peninsula; perturbations outside this region remain near zero,
confirming that F590 encodes a spatially selective heatwave signal.
The EU-mean $T_{2m}$ time series shifts upward coherently across all
five heatwave days (Figure~\ref{fig:steering}b), with the largest
response on 27 June 2019---the peak day where F590 has its highest
baseline activation.
 
The dose--response relationship is near-perfectly linear
(Figure~\ref{fig:steering}c): plotting $\Delta T_{2m}$ vs.\ $\alpha$
for each day yields $r^2 > 0.99$, implying that F590 encodes a
continuous heatwave intensity axis with no saturation at $\alpha = 2$.
Steering also perturbs $Z_{500}$ geopotential height in parallel with
$T_{2m}$ (Figure~\ref{fig:steering}d): the simultaneous response of
a surface variable and an upper-level variable is consistent with the
known dynamics of a blocking anticyclone, where upper-level ridging
and surface warming are tightly coupled.
MSLP over the European domain decreases monotonically with $\alpha$
(Figure~\ref{fig:steering}e), consistent with the surface thermal-low
signature that accompanies radiative heating under a blocking ridge.
 
\subsection{Mechanistic Circuit Analysis}

\begin{figure}[t]
  \centering
  \includegraphics[width=\linewidth]{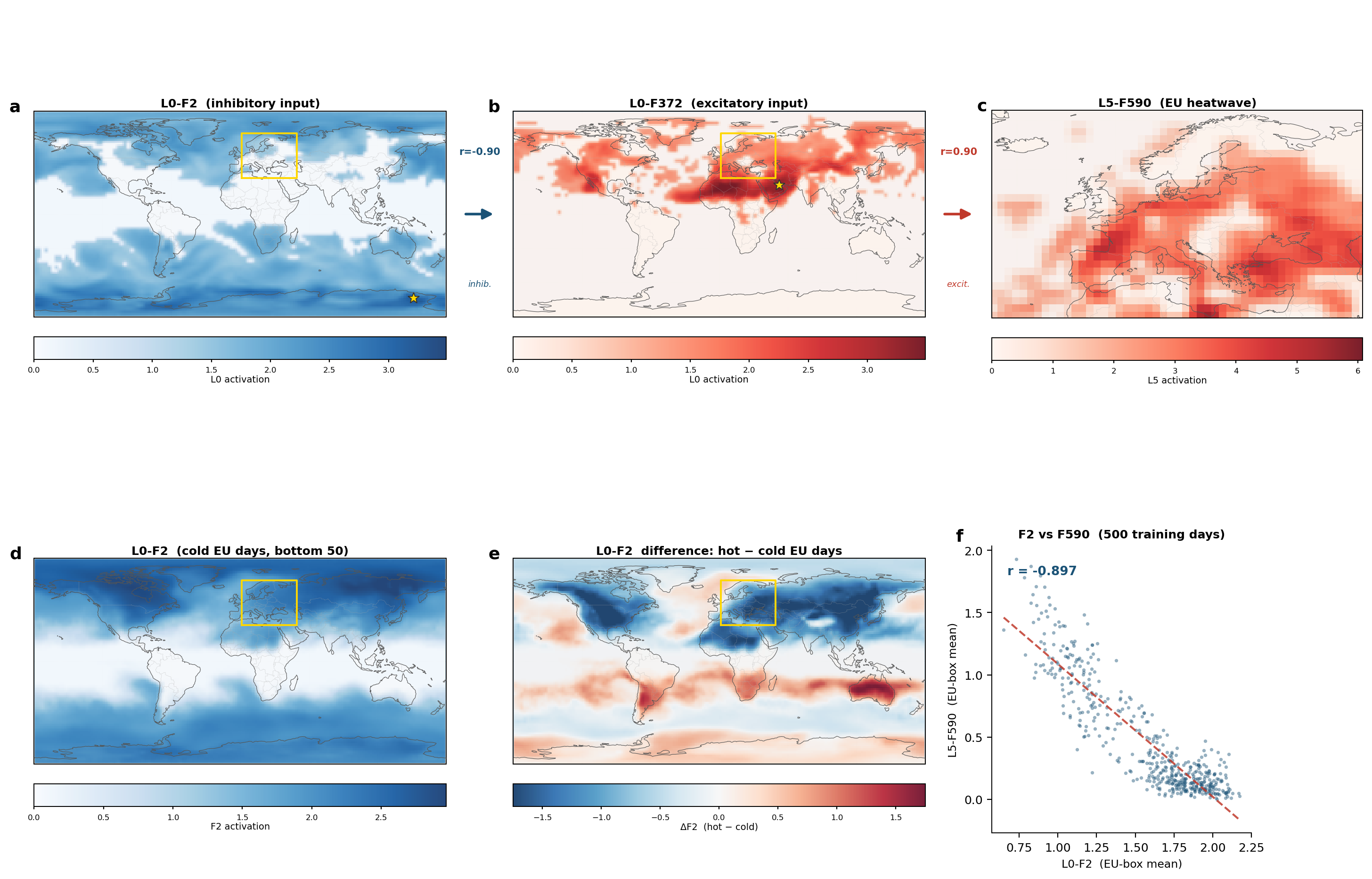}
  \caption{
    Push--pull circuit driving the Layer-5 heatwave feature F590.
    (a,d) Activation maps of the inhibitory feature L0-F2
    on hot and cold EU days, respectively; L0-F2 is suppressed on
    heatwave days.
    (b) Activation map of the excitatory feature L0-F372
    ($r = 0.90$ with F590); it activates broadly over Eurasia and
    is stronger on hot days.
    (e) Hot-minus-cold difference map for L0-F2, showing
    European suppression co-occurring with F590 activation.
    (f) Scatter of L0-F2 vs.\ L5-F590 activations over 500
    days ($r = -0.897$), confirming the inhibitory relationship.
  }
  \label{fig:circuit}
\end{figure}

Figure~\ref{fig:circuit} traces the Layer-0 antecedents of the
Layer-5 heatwave feature F590.
We identify the Layer-0 \KANSAE{} features most correlated with F590
over 500 training days and examine their spatial activation patterns.

Two Layer-0 features emerge as key circuit elements.
L0-F372 is excitatory: it activates broadly over the Eurasian
continent with a pattern strongly correlated with L5-F590
($r = 0.90$, Figure~\ref{fig:circuit}b), and its activation is
consistently stronger on hot EU days than cold days.
L0-F2 is inhibitory: it is globally active with a negative
correlation with F590 ($r = -0.897$, Figure~\ref{fig:circuit}f),
and is substantially stronger on cold EU days
(Figure~\ref{fig:circuit}d) than hot days
(Figure~\ref{fig:circuit}a).
The hot-minus-cold difference map (Figure~\ref{fig:circuit}e)
highlights the European region as the locus of L0-F2 suppression
that accompanies F590 activation.

This circuit suggests a push--pull mechanism: heatwave conditions
simultaneously excite a broad Eurasian pattern (L0-F372) and
suppress a global inhibitory feature (L0-F2), together driving the
sharp activation of L5-F590. This mirrors the push--pull circuits found in NLP transformers~\cite{olah2020zoom}, here manifesting as a physically
interpretable atmospheric pathway.

\subsection{Reconstruction Fidelity}

\begin{figure}[t]
  \centering
  \includegraphics[width=\linewidth]{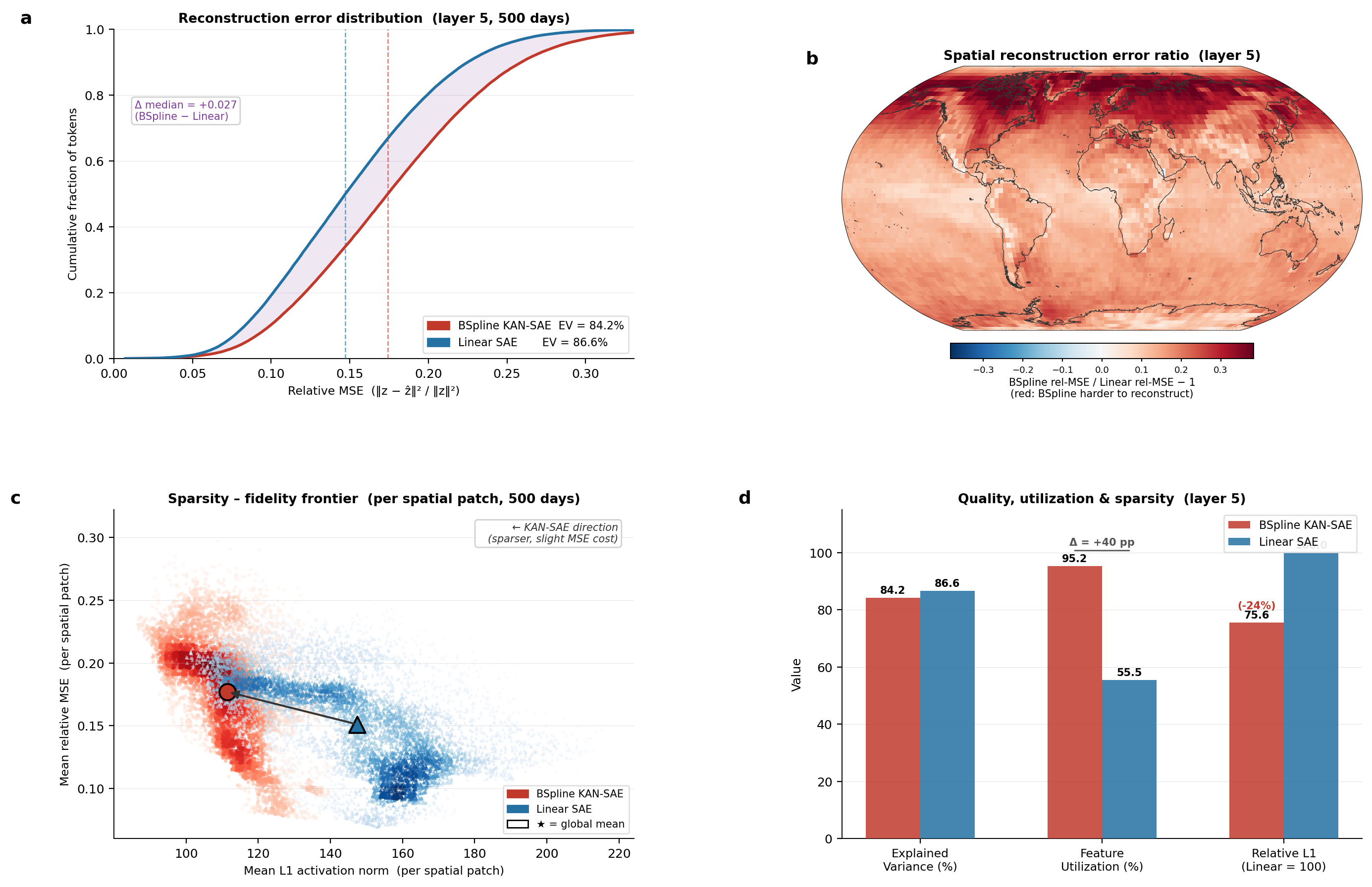}
  \caption{
    Reconstruction fidelity comparison between \KANSAE{} and \LinSAE{}
    at Layer 5.
    (a) CDF of per-token relative MSE: the two models achieve
    similar reconstruction error despite \KANSAE{} using 24\% lower
    mean $\ell_1$ norm.
    (b) Global spatial ratio of reconstruction MSE
    (\KANSAE{} / \LinSAE{}); values near 1 indicate comparable fidelity
    across most of the globe, with \KANSAE{} outperforming \LinSAE{}
    over the tropical Pacific.
    (c) Sparsity--fidelity scatter per spatial patch:
    \KANSAE{} (circles) achieves similar MSE at lower $\ell_1$,
    confirming more parsimonious encoding.
    (d) Aggregate summary metrics.
  }
  \label{fig:fidelity}
\end{figure}

Figure~\ref{fig:fidelity} examines reconstruction quality in detail.
The MSE CDF (Figure~\ref{fig:fidelity}a) places \KANSAE{}
(EV~84.2\%) slightly right of \LinSAE{} (EV~86.6\%), confirming
the modest 2.4-point aggregate gap reported in
Table~\ref{tab:feature_quality}.
The sparsity--fidelity scatter (Figure~\ref{fig:fidelity}c) reveals
that \KANSAE{} achieves comparable per-patch reconstruction error at
24\% lower mean $\ell_1$ norm, indicating more parsimonious encoding:
the same information is represented with fewer active features.

Spatially (Figure~\ref{fig:fidelity}b), the MSE ratio is near unity
across most of the globe; \KANSAE{} is harder to reconstruct only in
mid-latitude land regions, likely reflecting the high variability of
temperature extremes that \KANSAE{} encodes with fewer, more selective
features.
The summary bar chart (Figure~\ref{fig:fidelity}d) confirms the
three-way trade-off: \KANSAE{} trades ${\sim}2\%$ EV for a 72\%
gain in feature utilisation and a 24\% reduction in activation norm,
consistent with the additional per-feature parameters introduced by
the B-spline encoder.

\clearpage
\section{Conclusion}
\label{sec:conclusion}

We presented \KANSAE{}, a nonlinear sparse autoencoder that uses
learnable B-spline activation functions to decompose the internal
representations of a deep learning weather prediction model.
By relaxing the linear superposition assumption, \KANSAE{} discovers
significantly more alive features (95.2\% vs.\ 55.5\%) with lower
inter-feature redundancy (20\% reduction in median $|r|$), while
maintaining high reconstruction fidelity.
Case studies reveal physically interpretable features for European
heatwaves and tropical typhoons, validated through causal steering
experiments.
Circuit analysis shows that features become progressively more
event-specific with network depth, mirroring the hierarchical
processing observed in NLP transformer circuits.

Our analysis is currently restricted to Layer~5 of one model
(\Sonny{}) and two climate phenomena in the European and western
Pacific domains.
Generalisation to other layers, models, and phenomena remains to be
shown, and a principled criterion for the alive feature threshold
$\tau$ would improve reproducibility.
Future work could extend \KANSAE{} to other layers and models to
build a more complete mechanistic account of how deep learning weather
prediction models represent atmospheric dynamics, and investigate
whether the discovered features and circuits transfer across
architectures.

-----------------------------------------------------------------------
\bibliography{references}
\bibliographystyle{plainnat}

\end{document}